\pdfoutput=1

\documentclass[11pt]{article}

\usepackage[final]{acl}

\usepackage{times}
\usepackage{latexsym}

\usepackage[T1]{fontenc}

\usepackage[utf8]{inputenc}

\usepackage{microtype}

\usepackage{inconsolata}

\usepackage{graphicx}
\usepackage{booktabs} 
\usepackage{multirow}
\usepackage{makecell}
\usepackage{arydshln} 
\usepackage{tabularx}  
\usepackage{subcaption}
\usepackage{footmisc}  

\definecolor{takeaway-blue}{RGB}{218,227,243}
\definecolor{takeaway-title-blue}{RGB}{51,74,133}
\usepackage[most,skins,theorems]{tcolorbox}

\tcbset{
  aibox/.style={
    width=\linewidth,
    top=10pt,
    bottom=4pt,
    left=2pt,
    right=2pt,
    colback=takeaway-blue,
    colframe=black,
    colbacktitle=takeaway-title-blue,
    boxrule=0.5pt,
    enhanced,
    center,
    attach boxed title to top left={yshift=-0.1in,xshift=0.15in},
    boxed title style={boxrule=0pt,colframe=white,},
    breakable
  }
}
\newtcolorbox{AIbox}[2][]{aibox,title=#2,#1}

\title{Reasoning Does Not Necessarily Improve Role-Playing Ability}

\author{
\bf Xiachong Feng$^{\heartsuit}$$^{\spadesuit}$,
Longxu Dou$^{\spadesuit}$\thanks{Corresponding Authors.},
Lingpeng Kong$^{\heartsuit}$\footnotemark[1]\\
$^{\heartsuit}$The University of Hong Kong, $^{\spadesuit}$Sea AI Lab \\
{\tt fengxc@hku.hk, doulx@sea.com, lpk@cs.hku.hk}
}

\begin{document}
\maketitle
\begin{abstract}
The application of role-playing large language models (LLMs) is rapidly expanding in both academic and commercial domains, driving an increasing demand for high-precision role-playing models. 
Simultaneously, the rapid advancement of reasoning techniques has continuously pushed the performance boundaries of LLMs. 
This intersection of practical role-playing demands and evolving reasoning capabilities raises an important research question: “Can reasoning techniques enhance the role-playing capabilities of LLMs?” 
To address this, we conduct a comprehensive study using 6 role-playing benchmarks, 24 LLMs, and 3 distinct role-playing strategies, comparing the effectiveness of direct zero-shot role-playing, role-playing with Chain-of-Thought (CoT), and role-playing using reasoning-optimized LLMs. 
Our findings reveal that CoT may reduce role-playing performance, reasoning-optimized LLMs are unsuitable for role-playing, reasoning ability disrupts the role-playing scaling law, large models still lack proficiency in advanced role-playing, and Chinese role-playing performance surpasses English role-playing performance.
Furthermore, based on extensive experimental results, we propose two promising future research directions: Role-aware CoT for improving role-playing LLMs and Reinforcement Learning for role-playing LLMs, aiming to enhance the adaptability, consistency, and effectiveness of role-playing LLMs for both research and real-world applications\footnote{Our code is available at \href{https://github.com/xcfcode/rolecompass-acl}{xcfcode/rolecompass}.}.
\end{abstract}

\section{Introduction}

Large language models (LLMs), leveraging their strong foundational capabilities and advanced instruction-following abilities, have become the cornerstone of role-playing AI models~\citep{chen2024persona,tseng2024two}.
These role-playing LLMs present new opportunities in both research and commercial applications. 
In the academic domain, they offer novel possibilities for social simulation experiments~\citep{mou2024individual}. 
In industry, role-specific assistants, emotional companionship AI, and intelligent game NPCs~\citep{park2023generative,xu2024can} hold significant potential for further development and commercialization.
At the same time, reasoning capabilities in LLMs have been further enhanced through Chain-of-Thought (CoT) reasoning~\citep{wei2022chain} and reinforcement learning~\citep{guo2025deepseek}, enabling them to tackle increasingly complex problems.

\begin{table}[t]
\small
    \centering
    \begin{tabularx}{\columnwidth}{X}
        \toprule
        \textbf{Context}  \\
        Tom: Captain, your boat has been stolen. \\
        \midrule
        \textbf{Zero-shot role-playing with GPT-4-Turbo} \\
        Not stolen, mate—temporarily misplaced. And I always get my ship back, savvy? \\
        \hdashline
        \textbf{Role-playing with chain-of-thought using GPT-4-Turbo} \\
        \textcolor{gray}{[Reasoning process omitted]}  So the Pearl's decided to go on a little adventure without me—no matter, I’ll just steal her back. \\
        \hdashline
        \textbf{Role-playing with OpenAI o1-mini} \\
        \textcolor{gray}{[Reasoning process omitted]}  Stolen, you say, mate—why, I'd be far more worried for the soul foolish enough to filch the Pearl, savvy? \\
        \bottomrule
    \end{tabularx}
    \caption{Jack Sparrow role-playing example. \textit{(Jack Sparrow is a cunning and eccentric pirate captain known for his flamboyant, witty, and often rambling speech that conceals sharp intelligence beneath a layer of drunken charm).} It can be observed that direct zero-shot prompting of GPT yields better results, aligning more closely with Jack Sparrow's character traits.}
    \label{tab:intro}
\end{table}

This situation prompts us to explore a previously unanswered research question: “Can reasoning techniques enhance the role-playing capabilities of large language models?”
The answer to this question could provide valuable insights for the future application of reasoning techniques in role-playing LLMs, potentially paving a new path for their development and advancement.

To address this question, we carefully select six role-playing datasets and conduct experiments using 24 widely used open-source and proprietary models.
These models include API-based closed-source models such as GPT-4-Turbo~\citep{achiam2023gpt}, popular open-source models like Qwen2.5~\citep{yang2024qwen2}, and reinforcement learning-optimized reasoning models such as DeepSeek-R1~\citep{guo2025deepseek}.
All experiments are conducted using OpenCompass~\citep{2023opencompass} to ensure evaluation consistency, stability, and reproducibility. 
Specifically, we utilize six role-playing benchmarks: RoleBench~\citep{wang2023rolellm}, InCharacter~\citep{wang2024incharacter}, SocialBench~\citep{chen2024socialbench}, CharacterEval~\citep{tu2024charactereval}, HPD~\citep{chen2023hpd}, and CroSS-MR~\citep{yuan2024cross}. 
These benchmarks cover both multiple-choice and text generation tasks and include English and Chinese, making the evaluation linguistically diverse.
For evaluation, we employ both traditional metrics (e.g., Accuracy, ROUGE, and Exact Match) and LLM-as-a-Judge methods (e.g., prompt-based evaluation and reward model scoring), ensuring a broad assessment scope to enhance the robustness of our conclusions.
Furthermore, we define three role-playing approaches: direct zero-shot role-playing, role-playing with Chain-of-Thought (CoT), and role-playing using reasoning-optimized LLMs. 
This comprehensive setup allows us to systematically investigate the impact of reasoning techniques on role-playing performance.
Table~\ref{tab:intro} presents a specific example, demonstrating that the first setting, direct zero-shot role-playing, achieves the best performance.

Through extensive experiments, our findings reveal several key insights that can inform the future design of role-playing LLMs:
(1) CoT may reduce role-playing performance;
(2) Reasoning-optimized LLMs are unsuitable for role-playing;
(3) Reasoning ability disrupts the role-playing scaling law;
(4) The Qwen series is well-suited for role-playing;
(5) Chinese role-playing performance surpasses English role-playing performance;
(6) Large models still lack proficiency in advanced role-playing.

Our study makes three key contributions. First, through extensive experiments on six role-playing datasets and 24 models, we reveal that reasoning techniques do not necessarily enhance the role-playing capabilities of LLMs. 
Second, our results provide a comprehensive analysis of the current state and limitations of role-playing LLMs, leading to two promising research directions: Role-Aware CoT for Improving Role-Playing LLMs and Reinforcement Learning for Role-Playing LLMs. 
Third, to address the fragmented nature of prior work with inconsistent evaluation standards, we integrate all datasets and experimental methods into OpenCompass, enabling one-click execution and providing a standardized experimental framework and codebase for future research\footnote{All code will be open-sourced upon publication.}.

\section{Evaluation Framework}
The evaluation framework for LLM-based role-playing agents primarily consists of the following components: the LLM itself (\(\mathcal{M}\)), character profile (\(\mathcal{P}\)), evaluation task (\(\mathcal{T}\)), evaluation metric (\(\mathcal{E}\)), and, when applicable, a reference standard answer (\(\mathcal{A}\)).
In this study, we focus on whether reasoning techniques (\(\mathcal{R}\)) can enhance the role-playing capabilities of LLMs. 
Therefore, the overall evaluation process can be formalized as follows:

Given an LLM \(\mathcal{M}\), a predefined character profile \(\mathcal{P}\), an evaluation task \(\mathcal{T}\), and an evaluation metric \(\mathcal{E}\), the role-playing performance of \(\mathcal{M}\) is assessed by generating responses conditioned on \(\mathcal{P}\) and \(\mathcal{T}\). The generated outputs are then evaluated using \(\mathcal{E}\), which quantifies alignment with the intended role. 
Additionally, reasoning techniques \(\mathcal{R}\) are incorporated into \(\mathcal{M}\) to examine their impact on enhancing role-playing abilities.
Formally, the evaluation process can be expressed as:
$S = \mathcal{E}(\mathcal{M}(\mathcal{P}, \mathcal{T}, \mathcal{R}), \mathcal{A})$
where \( S \) represents the final performance score, capturing the effectiveness of \(\mathcal{M}\) in role-playing under the given conditions. 
A higher \( S \) indicates stronger role-playing capabilities.
If \(\mathcal{A}\) is unavailable or unnecessary, the evaluation metric \(\mathcal{E}\) can be adapted to rely on alternative assessment criteria, such as LLM-as-a-Judge.

\begin{table*}[t]
    \centering
    \resizebox{\textwidth}{!}{%
    \begin{tabular}{llllllll}
        \toprule
        \textbf{Dataset} & \textbf{Task Type} & \textbf{Evaluation Metric} & \textbf{Task} & \textbf{Language} & \textbf{\#Characters} & \textbf{\#Samples} \\
        \midrule
        \multirow{3}{*}{RoleBench~\citep{wang2023rolellm}} & \multirow{3}{*}{Generation} & \multirow{3}{*}{\makecell[l]{Automatic\\(ROUGE)}} & \multirow{2}{*}{General Question Answering} & EN & 95 & 32833 \\
        &  &  &  & ZH & 5 & 1690 \\
        \cmidrule{4-7}
        &  &  & Role-specific Question Answering & EN & 95 & 7534 \\
        \midrule
        \multirow{2}{*}{HPD~\citep{chen2023hpd}} 
        & \multirow{2}{*}{Generation} & \multirow{2}{*}{\makecell[l]{Automatic\\(ROUGE)}} & \multirow{2}{*}{Response Generation} & EN & 1 & 149 \\
        &  &  &  & ZH & 1 & 167 \\
        \midrule
        CharacterEval~\citep{tu2024charactereval} & Generation & \makecell[l]{LLM-as-a-Judge\\(Pretrained Reward Model)} & Response Generation & ZH & 77 & 4564 \\
        \midrule
        \textsc{CroSS-MR}~\citep{yuan2024cross} & Multiple-choice & \makecell[l]{Automatic\\(Accuracy)} & Motivation Recognition & EN & 126 & 445 \\
        \midrule
        \multirow{10}{*}{Socialbench~\citep{chen2024socialbench}} 
        & \multirow{6}{*}{Multiple-choice} & \multirow{6}{*}{\makecell[l]{Automatic\\(Accuracy)}} & \multirow{2}{*}{Role Knowledge Understanding} & EN & 23 & 988 \\
        &  &  &  & ZH & 15 & 405 \\
        \cmidrule{4-7}
        &  &  & \multirow{2}{*}{Behavioral Style Understanding} & EN & 8 & 740 \\
        &  &  &  & ZH & 8 & 323  \\
        \cmidrule{4-7}
        &  &  & \multirow{2}{*}{Social Preference Recognition} & EN & 79 & 731 \\
        &  &  &  & ZH & 101 & 1185 \\
        \cmidrule{2-7}
        & \multirow{4}{*}{Generation} & \multirow{4}{*}{\makecell[l]{Automatic\\(Exact Match)}} & \multirow{2}{*}{Long Conversation Memorization} & EN & 47 & 728 \\
        &  &  &  & ZH & 35 & 620 \\
        \cmidrule{4-7}
        &  &  & \multirow{2}{*}{Short Conversation Memorization} & EN & 46 & 463 \\
        &  &  &  & ZH & 27 &  310 \\
        \midrule
        \multirow{2}{*}{InCharacter~\citep{wang2024incharacter}} & \multirow{2}{*}{Generation} & \multirow{2}{*}{\makecell[l]{LLM-as-a-Judge\\(Prompting)}} & 16Personalities Identification & EN & 32 & 32 \\
        \cmidrule{4-7}
        &  &  & BFI Identification & EN & 32 & 32 \\
        \bottomrule
    \end{tabular}
    }
    \caption{Benchmark statistics.}
    \label{tab:datasets}
\end{table*}

\section{Experimental Setting}
In this section, we introduce the benchmarks, models, metrics, and reasoning methods used in our experiments. 
All relevant code has been integrated into the OpenCompass~\citep{2023opencompass} library to facilitate efficient replication by researchers.

\subsection{Role-playing Benchmarks}
We select six role-playing benchmarks to ensure the reliability of our experimental results, including:
\textbf{RoleBench}~\citep{wang2023rolellm}: Generates responses to both general and role-specific questions based on role information.
\textbf{HPD}~\citep{chen2023hpd}: Produces responses by leveraging the provided Harry Potter character information and related character details.
\textbf{CharacterEval}~\citep{tu2024charactereval}: Generates responses based on detailed character information, including background experiences and personality traits.
\textbf{CroSS-MR}~\citep{yuan2024cross}: Recognizes a role’s behavioral motivations based on the provided character profile.
\textbf{SocialBench}~\citep{chen2024socialbench}: Understands a role’s knowledge, behavioral style, social preferences, and fine-grained memory based on the given role information.
\textbf{InCharacter}~\citep{wang2024incharacter}: Assesses a character’s personality through questionnaire-based interviews\footnote{InCharacter benchmark comprises 14 types of assessment questionnaires. In this study, we select the most classic ones: 16Personalities and BFI for reporting results. Meanwhile, all other questionnaires are also implemented in the code.}.
Detailed benchmark statistics are provided in Table~\ref{tab:datasets}.

\subsection{Backbone LLMs}
Our experiments include a total of 24 models, comprising 2 closed-source models and 22 open-source models. 
These models originate from six different companies, including state-of-the-art models such as GPT-4-Turbo and DeepSeek-R1. 
The detailed list of models is provided in Appendix~\ref{app:sec:models}.

\subsection{Evaluation Metrics}
Our evaluation incorporates two types of metrics: \textbf{automated metrics} and \textbf{LLM-as-a-Judge}, both of which are widely recognized and scalable evaluation approaches.
Automated metrics include \textit{ROUGE}, \textit{Accuracy}, and \textit{Exact Match}, which are well-established and widely accepted in the field as classical evaluation benchmarks.
LLM-as-a-Judge further encompasses two methodologies: the \textit{Pretrained Reward Model} and \textit{Prompting-based Scoring}. 
The Pretrained Reward Model is trained on annotated datasets and assigns scores to specified dimensions during evaluation. 
In contrast, Prompting-based Scoring leverages prompt engineering techniques and the extensive knowledge base of LLMs to dynamically and adaptively generate relevant scores, providing a more flexible and responsive evaluation mechanism.
In Appendix~\ref{app:sec:metrics}, we provide a detailed explanation of the evaluation metrics used for each benchmark.

\begin{figure*}[t]
    \centering
    \includegraphics[scale=0.41]{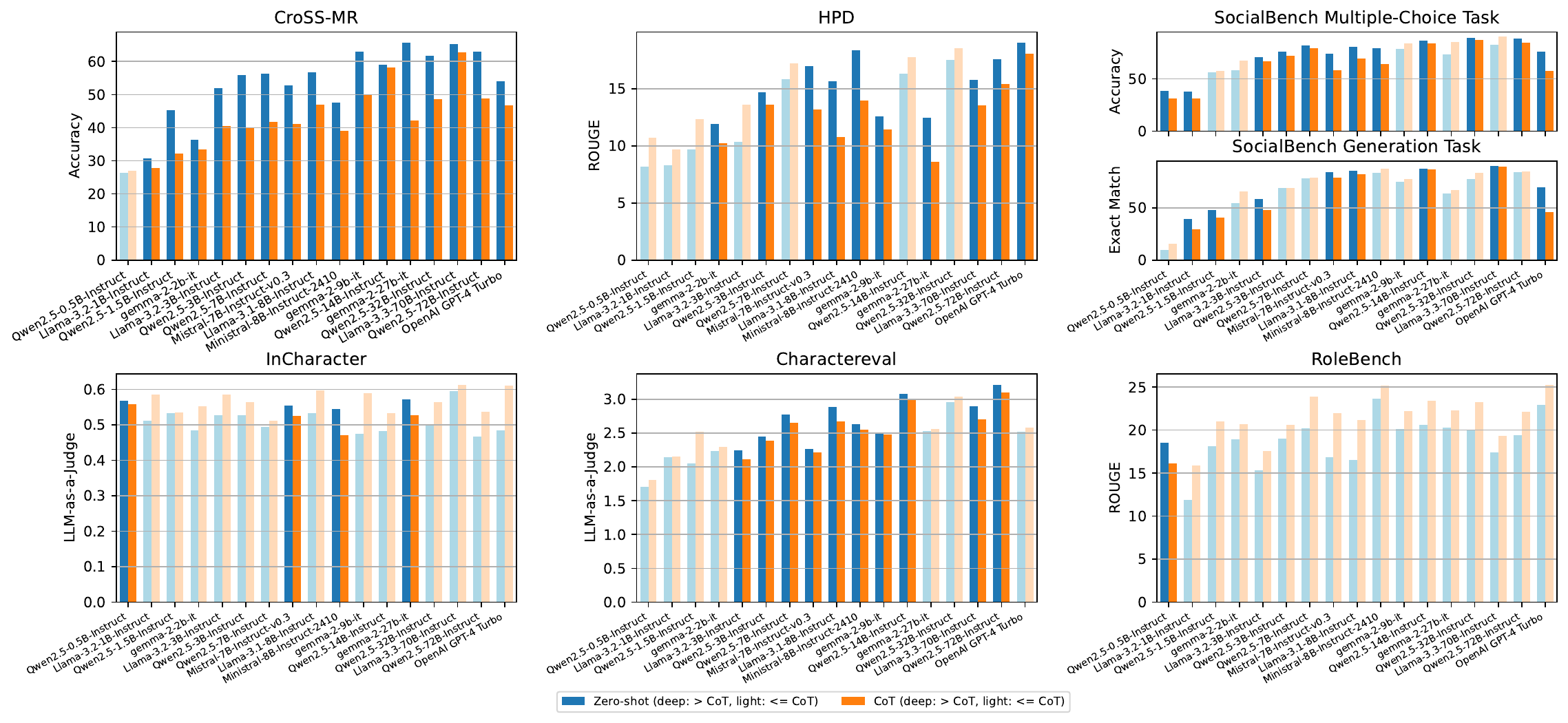}
    \caption{Performance comparison of 17 Models using two role-playing methods across six benchmarks. The horizontal axis ranks models in descending order of scale, while the vertical axis represents the unique metric for each dataset. Notably, darker colors indicate that zero-shot role-playing outperforms CoT role-playing, whereas lighter colors signify that zero-shot role-playing underperforms compared to CoT role-playing.}
    \label{fig:zero_vs_cot}
\end{figure*}

\subsection{Role-playing Methods}
In this study, we employ three approaches to guide LLMs in performing role-playing tasks:
\(\mathcal{R}_1\): Direct zero-shot role-playing using an LLM, where the model generates responses without reasoning steps.
\(\mathcal{R}_2\): Role-playing with chain-of-thought (CoT) reasoning, where an LLM explicitly engages in step-by-step reasoning before executing the role-playing task.
\(\mathcal{R}_3\): Role-playing using reasoning-optimized LLMs, such as \textit{QwQ-32B-Preview} and \textit{DeepSeek-R1}, which autonomously engage in deep reasoning before generating responses.

\section{Results and Findings}
In this section, we present our experimental analysis and key findings.

\subsection{CoT May Reduce Role-Playing Performance}

To investigate whether reasoning techniques enhance the role-playing capabilities of LLMs, we conduct extensive experiments. 
Specifically, we select 17 models and evaluate their role-playing performance using both zero-shot and chain-of-thought (CoT) approaches across six standardized and widely used benchmarks.
All experimental results are presented in the Appendix~\ref{app:sec:results}. 
Figure~\ref{fig:zero_vs_cot} provides an aggregated overview of the results. 
Specifically, for each model, its final performance on a given benchmark is computed as the average of its performance across all sub-datasets.

As shown in Figure~\ref{fig:zero_vs_cot}, employing CoT reasoning is more likely to degrade role-playing performance on four benchmarks: CroSS-MR, HPD, SocialBench, and CharacterEval. 
In contrast, on InCharacter and RoleBench, CoT reasoning enhances the role-playing capabilities of LLMs.
To further understand why CoT reduces the role-playing capabilities of LLMs, we select Qwen2.5-7B-Instruct as the experimental model. 
We then sample 50 test cases from each of the six benchmarks where CoT performance is lower than Zero-shot performance for detailed analysis.

Our findings indicate that the primary reasons for CoT-induced degradation in role-playing are:
(1) “Attention Diversion”: The model must simultaneously engage in reasoning and role-playing modes, which dilutes its focus on the role-playing task.
(2) “Linguistic Style Drift”: Reasoning responses tend to be structured, logical, and formal, whereas effective role-playing requires a vivid, expressive, and character-consistent linguistic style. 

\begin{AIbox}
{Finding 1:}
CoT may reduce the role-playing capabilities of LLMs, primarily due to attention diversion and linguistic style drift induced by the reasoning process.
\end{AIbox}

\begin{figure*}[t]
    \centering
    \includegraphics[scale=0.38]{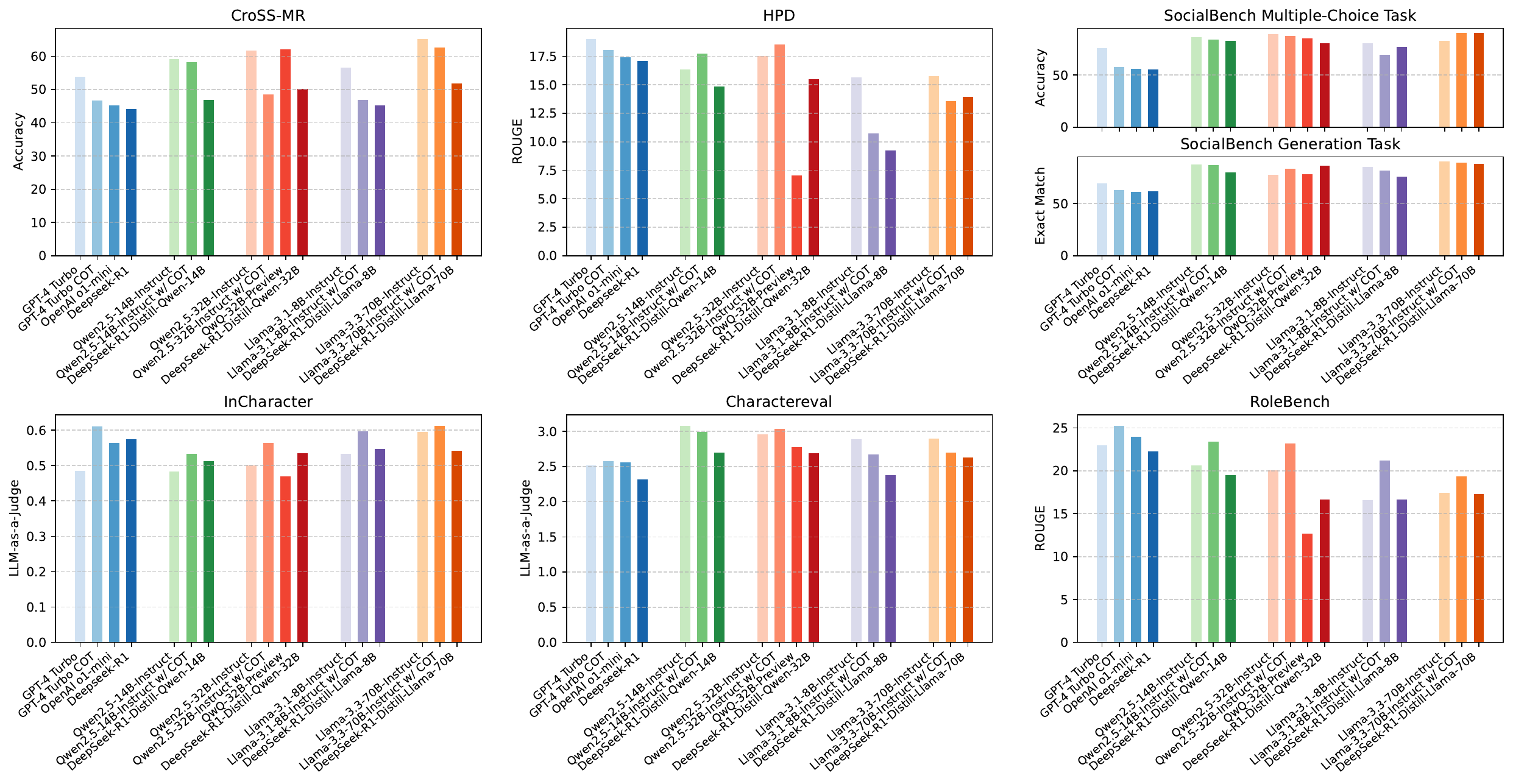}
    \caption{Experimental results of various models across 6 benchmarks. Models of similar sizes are represented using the same color scheme, with each employing different types of reasoning techniques. The vertical axis denotes the evaluation metrics specific to each dataset.}
    \label{fig:o1_vs}
\end{figure*}

\subsection{Reasoning-optimized LLMs Are Unsuitable for Role-Playing}
The most advanced models available today are undoubtedly reasoning-optimized models, including OpenAI's o series, DeepSeek's R1 model, and various distilled versions derived from DeepSeek-R1\footnote{\url{https://github.com/deepseek-ai/DeepSeek-R1}}.
Compared to CoT reasoning, these models leverage pretraining and reinforcement learning techniques to cultivate intrinsic reasoning capabilities, making them inherently more adept at reflection, verification, and other cognitive processes.

To investigate whether reasoning-optimized models are better suited for role-playing tasks, we conduct experiments using OpenAI o1-mini, QwQ-32B-Preview, DeepSeek-R1, and its various distilled versions. 
The experimental results are presented in Figure~\ref{fig:o1_vs}.

It is evident that reasoning-optimized LLMs generally perform poorly and are not well-suited for role-playing tasks, even for state-of-the-art models such as OpenAI o1-Mini and DeepSeek-R1. 
Furthermore, we observe that models refined through reasoning distillation exhibit even worse role-playing performance compared to their original counterparts. 
This finding aligns with the conclusion drawn in the previous subsection: enhancing rational reasoning capabilities tends to undermine the emotional and intuitive aspects essential for effective role-playing.
We provide more discussion in Appendix~\ref{app:sec:rllimit}.

\begin{AIbox}
{Finding 2:}
Reasoning-optimized LLMs are less suitable for role-playing tasks.
\end{AIbox}

\subsection{Reasoning Ability Disrupts the Role-Playing Scaling Law}

\begin{figure*}[!htb]
    \centering
    \begin{subfigure}[b]{\textwidth}
        \centering
        \includegraphics[width=\textwidth]{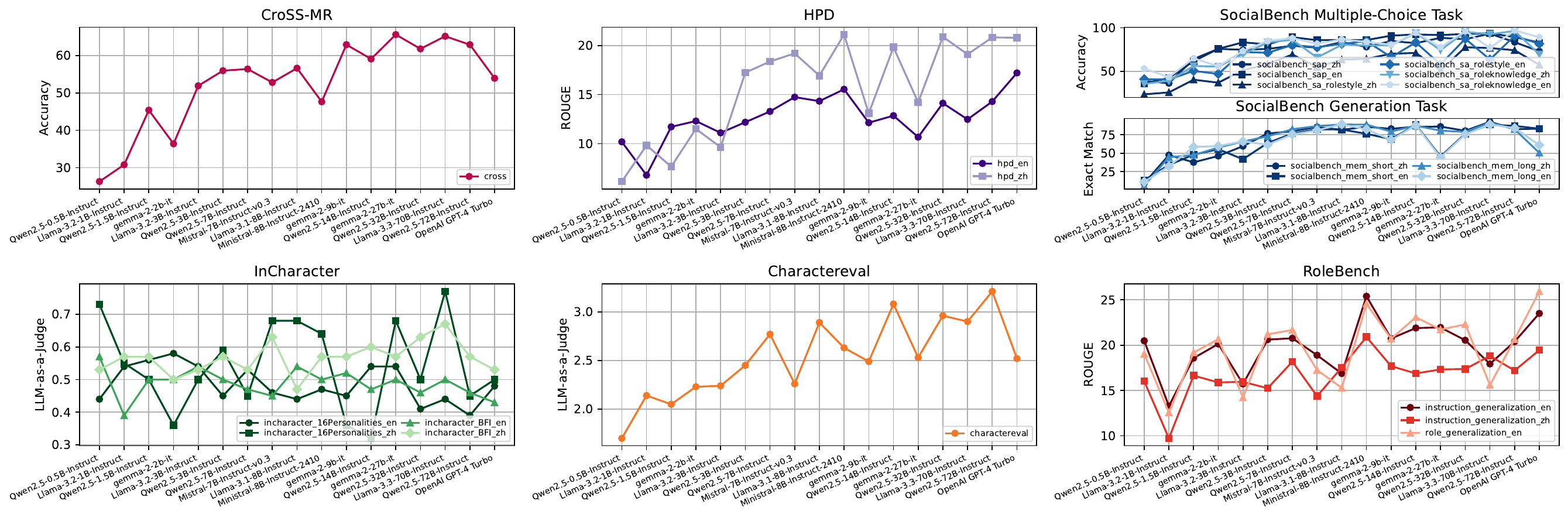}
        \caption{Results of direct zero-shot role-playing.}
        \label{fig:zero_shot}
    \end{subfigure}
    
    \begin{subfigure}[b]{\textwidth}
        \centering
        \includegraphics[width=\textwidth]{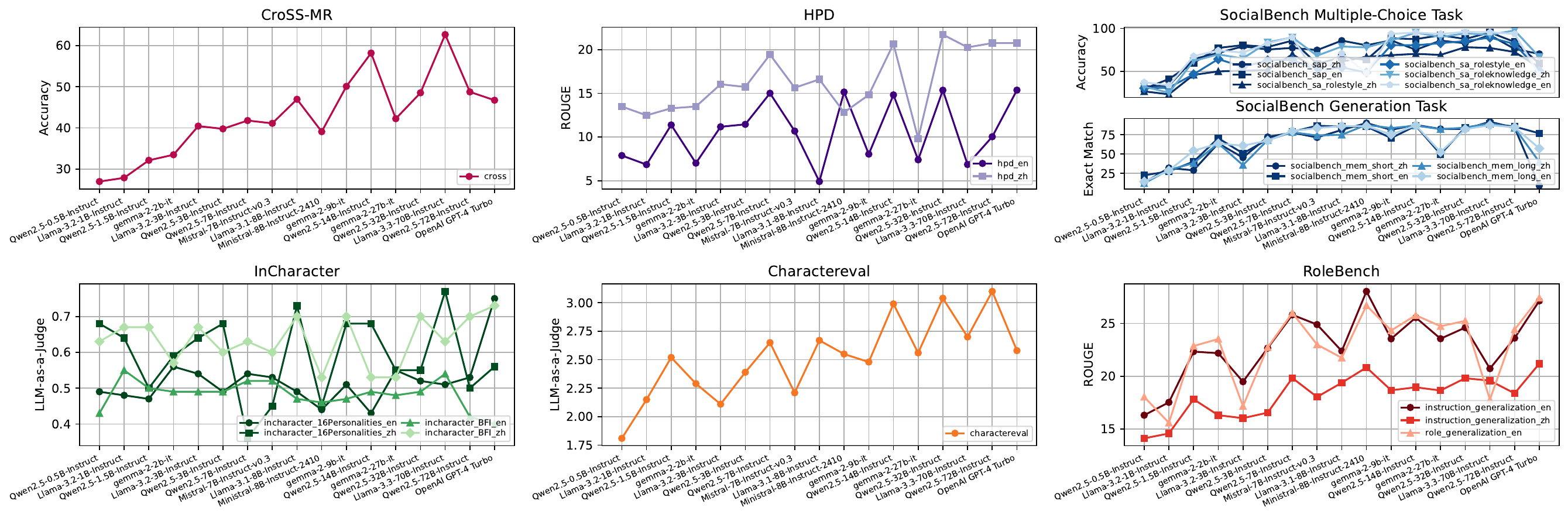}
        \caption{Results of role-playing with chain-of-thought (CoT) reasoning.}
        \label{fig:cot}
    \end{subfigure}

    \begin{subfigure}[b]{\textwidth}
        \centering
        \includegraphics[width=\textwidth]{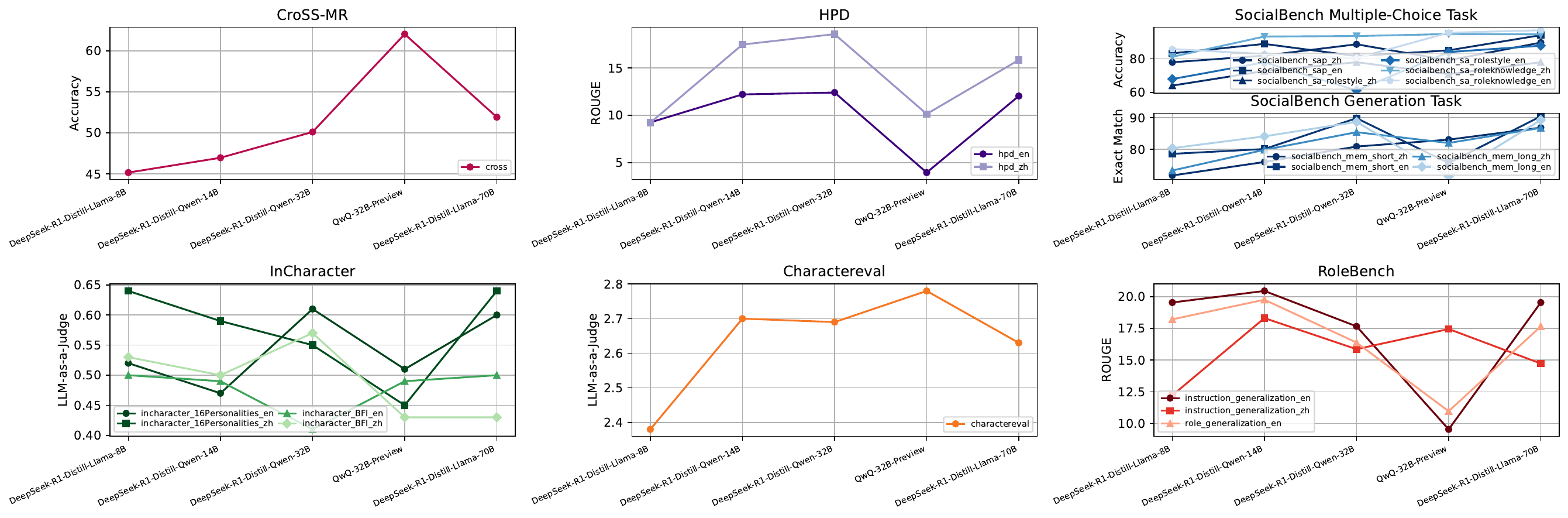}
        \caption{Results of role-playing using reasoning-optimized LLMs.}
        \label{fig:o1}
    \end{subfigure}

    \caption{Performance comparison of different models across six benchmarks. The horizontal axis represents model size, arranged from smallest to largest, while the vertical axis denotes benchmark-specific evaluation metrics, where higher values indicate better role-playing performance. Within each benchmark, different color gradients represent the performance curves for its respective sub-datasets.}
    \label{fig:all}
\end{figure*}

Figure~\ref{fig:all} presents the results of our three experimental settings: direct zero-shot role-playing, role-playing with Chain-of-Thought (CoT), and role-playing using reasoning-optimized LLMs. 
First, we observe that, with the exception of the InCharacter benchmark, the other five benchmarks generally follow the scaling law, where larger models exhibit stronger role-playing capabilities. 
However, we also find that the role-playing scaling law is not particularly pronounced—the performance gains from increasing model size remain relatively modest and inconsistent across benchmarks. 
Furthermore, introducing reasoning capabilities, whether through CoT or reasoning-optimized LLMs, further weakens the benefits of scaling, leading to more pronounced fluctuations, increased instability, and greater variability in model performance across different role-playing tasks and datasets, making the scaling trend less predictable and consistent.

\begin{AIbox}
{Finding 3:}
The role-playing scaling law exists but is not pronounced, and the introduction of reasoning capabilities disrupts this scaling law.
\end{AIbox}

\subsection{The Qwen Series Is Well-Suited for Role-Playing Tasks}
To provide guidance on model selection for future role-playing tasks, we conduct a comprehensive and systematic comparative analysis across different model scales (e.g., 1B, 3B, 7B, 14B, 32B, 72B).
The results are shown in Figure~\ref{fig:all}.
Our findings indicate that the Qwen2.5 series consistently outperforms the Llama, Gemma, and Mistral series across various size ranges in terms of role-playing accuracy, persona consistency, linguistic expressiveness, and contextual coherence. 
Notably, we recommend Qwen2.5-7B-Instruct as the most cost-effective and well-balanced model for role-playing applications, as it offers a strong trade-off between performance, computational efficiency, and adaptability across diverse role-playing scenarios.

\begin{AIbox}
{Finding 4:}
The Qwen2.5 series excels in role-playing tasks, with Qwen2.5-7B-Instruct being the most cost-effective model.
\end{AIbox}

\begin{figure*}[t]
    \centering
    \includegraphics[scale=0.34]{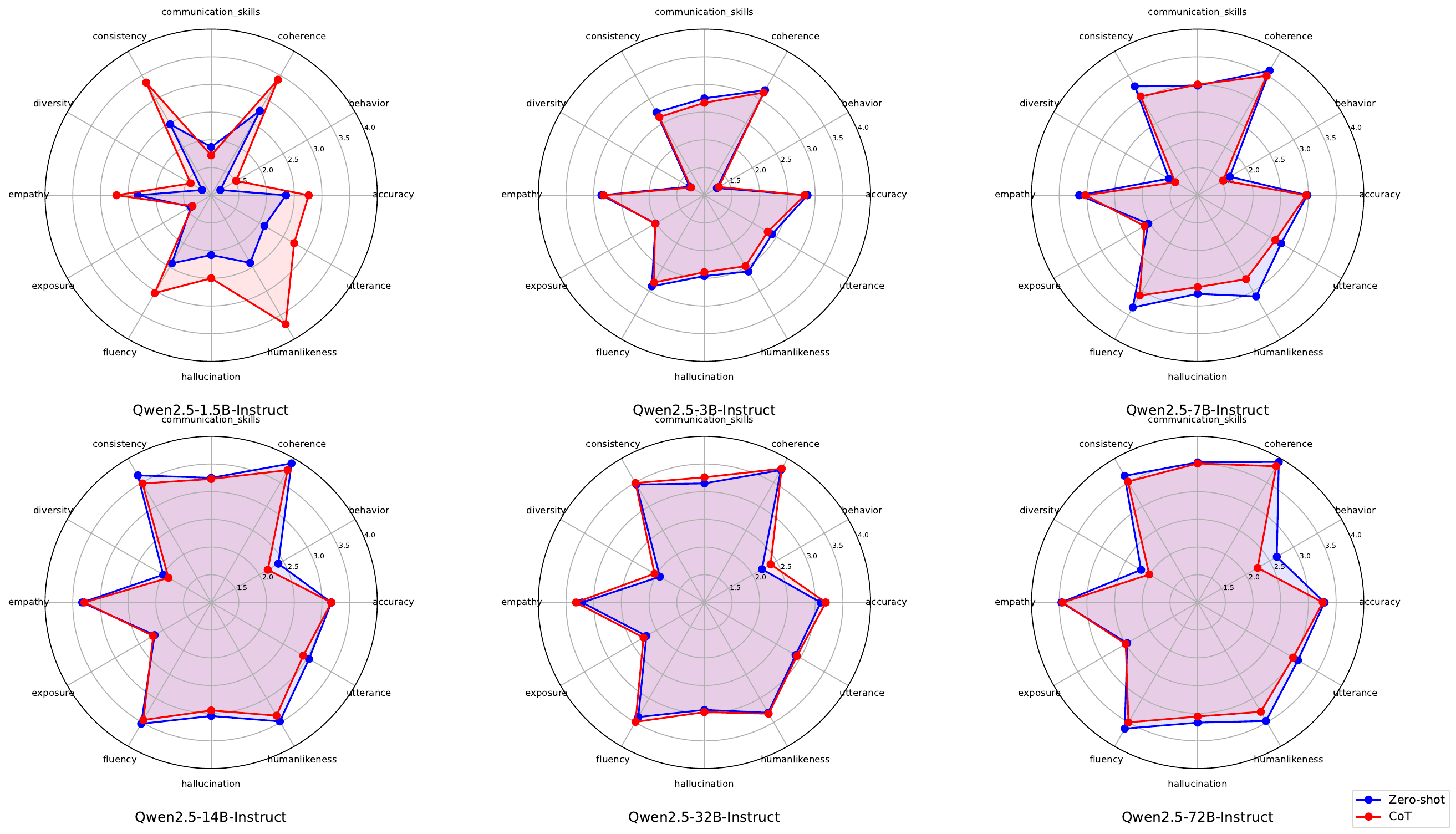}
    \caption{Fine-grained performance of the Qwen2.5 series on the CharacterEval benchmark. The radar chart illustrates multiple evaluation dimensions, with metrics computed using a pretrained reward model. Higher scores indicate stronger capabilities.}
    \label{fig:charactereval_radar}
\end{figure*}

\subsection{Chinese Role-Playing Performance Surpasses English Role-Playing Performance}

The role-playing benchmarks HPD, SocialBench, InCharacter, and RoleBench provide both Chinese and English versions, enabling a direct comparison of multilingual role-playing performance. 
Interestingly, an analysis of Figure~\ref{fig:all} reveals that current LLMs exhibit stronger role-playing capabilities in Chinese than in English across multiple benchmarks and evaluation metrics. 
This observation contradicts the common conclusion that LLMs generally have a stronger foundation in English compared to other languages, particularly in reasoning and knowledge-intensive tasks.
We hypothesize that this phenomenon may be attributed to the following factor:
Due to the extensive training on English data, models have to some extent internalized generalized character information. 
When performing precise role-playing tasks, this internalized generalization can interfere with context-sensitive role-playing, leading to performance degradation.

\begin{AIbox}
{Finding 5:}
Large models exhibit superior role-playing capabilities in Chinese compared to English to some extent.
\end{AIbox}

\subsection{Large Models Still Lack Proficiency in Advanced Role-Playing}
To further investigate the limitations of current LLM-based role-playing models, we conduct an analysis using the CharacterEval benchmark.
Specifically, CharacterEval provides a pretrained reward model that assigns scores across 12 role-playing evaluation dimensions\footnote{\url{https://huggingface.co/morecry/BaichuanCharRM}}. 
We analyze the performance of the Qwen2.5 series, with the results presented in Figure~\ref{fig:charactereval_radar}.
All detailed results are presented in Appendix~\ref{app:sec:charactereval}.
First, consistent with our previous findings, direct zero-shot role-playing outperforms other approaches in most cases.
Furthermore, a fine-grained comparison of evaluation metrics reveals that current LLMs still exhibit deficiencies in advanced role-playing dimensions, such as character knowledge exposition, personality expression, and linguistic diversity. 
These aspects require further enhancement in future research to improve the overall role-playing capability of LLMs.

\begin{AIbox}
{Finding 6:}
Current LLMs still underperform in advanced role-playing capabilities, such as knowledge exposition, personality expression, and diversity.
\end{AIbox}

\section{Related Works}

In recent years, large language models (LLMs) have revolutionized the paradigm of artificial intelligence, achieving human-like performance in mathematics~\citep{liu2023mathematical,shao2024deepseekmath}, coding~\citep{luo2023wizardcoder,roziere2023code,guo2024deepseek,zhu2024deepseek}, embodied intelligence~\citep{wang2023voyager,liu2024aligning}, game intelligence~\citep{hu2024survey,feng2024survey}, and advanced reasoning~\citep{chu2023survey,xu2025towards}. 
The comprehensive enhancement of foundational capabilities in these models has opened new opportunities for the development of role-playing agents~\citep{chen2024persona}, resulting in a proliferation of such applications based on LLMs.
On one hand, this has catalyzed innovations in fields such as psychological counseling, anthropomorphic companionship, and game NPCs~\citep{tseng2024two}. 
On the other hand, it has provided fresh avenues for advancing social simulations~\citep{mou2024individual}. 
Concurrently, reasoning techniques in LLMs have flourished, with methods like chain-of-thought reasoning~\citep{wei2022chain}, tree-of-thought reasoning~\citep{yao2024tree}, and o1-style reasoning~\citep{jaech2024openai} progressively maximizing the potential of these models, particularly achieving remarkable outcomes in tasks involving mathematics and coding~\citep{guo2025deepseek,team2025kimi}.
While prior research has predominantly focused on role-playing applications and their evaluation, this study is positioned at the intersection of LLM-based role-playing agents and reasoning techniques. 
It seeks to address the scientific question of whether reasoning techniques enhance the role-playing capabilities of LLMs. 
By doing so, this work aims to provide guidance for developing more realistic and reliable role-playing agents powered by LLMs.
\section{Future Directions}
Based on our experimental findings, we propose two potential research directions that integrate role-playing and reasoning techniques for future exploration.

\noindent \paragraph{Role-aware Chain-of-Thought (CoT) for Improving Role-playing LLMs.}
One promising direction for enhancing the role-playing ability of LLMs is Role-aware CoT reasoning. 
While standard CoT enables step-by-step logical inference, it often disregards the persona-specific constraints that are essential for consistent role-playing. 
A role-aware CoT approach would integrate persona attributes, narrative constraints, and character-specific perspectives into the reasoning process, ensuring that logical deductions align with the character's predefined traits. 
For instance, a historical figure simulated in an LLM should reason within the knowledge and biases of their era rather than applying contemporary logic. 
This requires incorporating dynamic memory structures that retain persona information throughout multi-turn interactions, mitigating the risk of breaking character or adopting inconsistent reasoning patterns.

\noindent\paragraph{Reinforcement Learning for Role-playing LLMs.}

DeepSeek-R1 has demonstrated that by defining precise rule-based rewards, reinforcement learning alone can induce emergent reasoning and cognitive capabilities. This suggests an important research direction: investigating whether carefully designed role-playing task rewards can enable models to autonomously develop intrinsic, role-specific reasoning and thinking abilities, thereby enhancing their role-playing performance.
A critical consideration in reward design is ensuring that it simultaneously accounts for the accuracy of the role-playing task while effectively guiding the model to develop role-specific reasoning abilities. The reward function should prevent the model from relying on shortcuts to generate final responses, instead encouraging it to engage in authentic character-driven reasoning and decision-making.
\section{Conclusion}
This study aims to address the research question: “Can reasoning techniques enhance the role-playing capabilities of large language models (LLMs)?” 
To this end, we conduct extensive experiments using 6 role-playing benchmarks, 24 LLMs, and 3 distinct role-playing methods. 
Our experimental results lead to the following key findings: CoT may reduce role-playing performance, reasoning-optimized LLMs are unsuitable for role-playing, reasoning ability disrupts the role-playing scaling law, the Qwen series is well-suited for role-playing tasks, Chinese role-playing performance surpasses English role-playing performance, and large models still lack proficiency in advanced role-playing.
All code is integrated into OpenCompass, ensuring reproducibility and facilitating further research. 
We hope that our findings provide new perspectives for future studies on role-playing LLMs.

\clearpage
\newpage

\section*{Limitations}
Although we have made every effort to comprehensively cover the most commonly used models and role-playing benchmarks, our selection may still be incomplete. 
Certain niche models or specialized benchmarks may not have been included, which could introduce potential biases in our findings. 
As a result, while our conclusions provide valuable insights into the role-playing capabilities of large language models, their robustness could be further enhanced through broader model coverage, additional benchmark evaluations, and more extensive experimental validations. 
Future studies should explore a wider range of datasets and model architectures to ensure greater generalizability and reliability of the findings.

\subsubsection*{Acknowledgments}
This work was supported by Hong Kong Innovation and Technology Support Programme Platform Research Project fund (ITS/269/22FP).

\bibliography{custom}

\appendix

\section{Evaluated Models}\label{app:sec:models}
Table~\ref{tab:models} lists all the large language models (LLMs) used in this study.

\begin{table}[t]
    \centering
    \begin{tabular}{@{}ll@{}}
    \toprule
    \textbf{Team} & \textbf{Model} \\ 
    \midrule
    \multicolumn{2}{c}{\textbf{Closed-source}} \\ 
    \midrule
    \multirow{2}{*}{OpenAI} & GPT-4 Turbo \\
     & OpenAI o1-mini \\
    \midrule
    \multicolumn{2}{l}{\textbf{Open-source}} \\ 
    \midrule
    \multirow{8}{*}{Qwen} & Qwen2.5-0.5B-Instruct \\
     & Qwen2.5-1.5B-Instruct \\
     & Qwen2.5-3B-Instruct \\
     & Qwen2.5-7B-Instruct \\
     & Qwen2.5-14B-Instruct \\
     & Qwen2.5-32B-Instruct \\
     & Qwen2.5-72B-Instruct \\
     & QwQ-32B-Preview \\
    \hdashline
    \multirow{3}{*}{Google} & gemma-2-2b-it \\
     & gemma-2-9b-it \\
     & gemma-2-27b-it \\
    \hdashline
    \multirow{4}{*}{Meta} & Llama-3.2-1B-Instruct \\
     & Llama-3.2-3B-Instruct \\
     & Llama-3.1-8B-Instruct \\
     & Llama-3.3-70B-Instruct \\
    \hdashline
    \multirow{2}{*}{Mistral AI} & Ministral-8B-Instruct-2410 \\
     & Mistral-7B-Instruct-v0.3 \\
    \hdashline
    \multirow{5}{*}{Deepseek} & DeepSeek-R1 \\
     & DeepSeek-R1-Distill-Llama-8B \\
     & DeepSeek-R1-Distill-Qwen-14B \\
     & DeepSeek-R1-Distill-Qwen-32B \\
     & DeepSeek-R1-Distill-Llama-70B \\
    \bottomrule
    \end{tabular}
\caption{List of models evaluated in the experiment.}
\label{tab:models}
\end{table}

\begin{table*}[ht]
\centering
\resizebox{\textwidth}{!}{%
\begin{tabular}{llcccccccc}
\toprule
\textbf{Dataset} & \textbf{Sub Dataset} & \textbf{\makecell{OpenAI \\ GPT-4 Turbo}} & \textbf{\makecell{Qwen2.5-0.5B\\-Instruct}} & \textbf{\makecell{Qwen2.5-1.5B\\-Instruct}} & \textbf{\makecell{Qwen2.5\\-3B-Instruct}} & \textbf{\makecell{Qwen2.5-7B\\-Instruct}} & \textbf{\makecell{Qwen2.5-14B\\-Instruct}} & \textbf{\makecell{Qwen2.5-32B\\-Instruct}} & \textbf{\makecell{Qwen2.5-72B\\-Instruct}} \\
\midrule
Cross & cross & 53.93 & 26.29 & 45.39 & 55.96 & 56.40 & 59.10 & 61.80 & 62.92 \\
\midrule
\multirow{2}{*}{HPD} & hpd\_en & 17.21 & 10.18 & 11.71 & 12.18 & 13.28 & 12.85 & 14.12 & 14.29 \\
 & hpd\_zh & 20.80 & 6.13 & 7.65 & 17.24 & 18.37 & 19.83 & 20.90 & 20.83 \\
 \midrule
\multirow{10}{*}{Socialbench} & socialbench\_sap\_zh & 73.08 & 39.04 & 64.22 & 75.78 & 79.41 & 82.36 & 86.92 & 83.54 \\
 & socialbench\_sap\_en & 83.04 & 36.99 & 61.70 & 80.98 & 89.19 & 92.34 & 92.89 & 89.33 \\
 & socialbench\_sa\_rolestyle\_zh & 57.59 & 23.84 & 40.56 & 58.51 & 69.35 & 71.21 & 77.71 & 74.30 \\
 & socialbench\_sa\_rolestyle\_en & 81.62 & 44.96 & 50.68 & 71.35 & 80.27 & 83.24 & 87.43 & 90.81 \\
 & socialbench\_sa\_roleknowledge\_zh & 89.07 & 53.09 & 65.08 & 85.02 & 88.16 & 94.03 & 96.26 & 96.30 \\
 & socialbench\_sa\_roleknowledge\_en & 83.13 & 44.03 & 37.63 & 76.63 & 79.62 & 85.91 & 80.27 & 96.96 \\
 & socialbench\_mem\_short\_zh & 83.16 & 13.35 & 48.03 & 63.83 & 76.50 & 88.41 & 76.33 & 82.53 \\
 & socialbench\_mem\_short\_en & 83.13 & 16.35 & 48.40 & 72.78 & 76.82 & 80.24 & 87.21 & 86.90 \\
 & socialbench\_mem\_long\_zh & 60.94 & 11.20 & 58.46 & 61.46 & 75.30 & 87.78 & 75.52 & 82.32 \\
 & socialbench\_mem\_long\_en & 60.91 & 14.01 & 58.46 & 61.46 & 75.30 & 87.78 & 75.52 & 83.92 \\
 \midrule
\multirow{4}{*}{InCharacter} & incharacter\_16Personalities\_en & 0.48 & 0.44 & 0.56 & 0.45 & 0.53 & 0.54 & 0.41 & 0.39 \\
 & incharacter\_16Personalities\_zh & 0.50 & 0.43 & 0.50 & 0.39 & 0.36 & 0.68 & 0.32 & 0.45 \\
 & incharacter\_BFI\_en & 0.50 & 0.50 & 0.50 & 0.57 & 0.47 & 0.47 & 0.53 & 0.46 \\
 & incharacter\_BFI\_zh & 0.53 & 0.57 & 0.57 & 0.53 & 0.60 & 0.60 & 0.63 & 0.57 \\
 \midrule
 Charactereval & charactereval & 2.52 & 1.70 & 2.05 & 2.45 & 2.77 & 3.08 & 2.96 & 3.21   \\
  \midrule
\multirow{3}{*}{RoleBench} & instruction\_generalization\_en & 18.57 & 20.41 & 20.67 & 20.73 & 21.15 & 21.89 & 13.28 & 3.21 \\
 & instruction\_generalization\_zh & 19.45 & 16.63 & 15.25 & 18.18 & 16.88 & 16.68 & 17.36 & 20.39 \\
 & role\_generalization\_en & 25.93 & 19.04 & 19.18 & 21.19 & 21.68 & 23.08 & 22.29 & 17.20 \\
\bottomrule
\end{tabular}%
}
\caption{Direct zero-shot role-playing performance of different LLMs on various role-playing benchmarks (part1).}
\label{tab:zero_shot_performance_part_1}
\end{table*}

\begin{table*}[ht]
\centering
\resizebox{\textwidth}{!}{%
\begin{tabular}{llccccccccc}
\toprule
\textbf{Dataset} & \textbf{Sub Dataset} & \textbf{\makecell{Gemma-2\\-2B-it}} & \textbf{\makecell{Gemma-2\\-9B-it}} & \textbf{\makecell{Gemma-2\\-27B-it}} & \textbf{\makecell{Llama-3.2-1B\\-Instruct}} & \textbf{\makecell{Llama-3.2-3B\\-Instruct}} & \textbf{\makecell{Llama-3.1-8B\\-Instruct}} & \textbf{\makecell{Llama-3.3-70B\\-Instruct}} & \textbf{\makecell{Mistral-8B\\-Instruct-2410}} & \textbf{\makecell{Mistral-7B\\-Instruct-v0.3}} \\
\midrule
Cross & cross & 36.40 & 62.92 & 65.62 & 30.79 & 51.91 & 56.63 & 65.17 & 47.64 & 52.81 \\
\midrule
\multirow{2}{*}{HPD} & hpd\_en & 12.30 & 12.13 & 10.66 & 6.78 & 11.10 & 14.33 & 12.47 & 15.54 & 14.74 \\
 & hpd\_zh & 11.50 & 13.08 & 14.20 & 9.82 & 9.63 & 16.93 & 19.10 & 21.12 & 19.20 \\
 \midrule
\multirow{10}{*}{SocialBench} & socialbench\_sap\_zh & 75.86 & 84.81 & 88.52 & 36.54 & 74.77 & 83.54 & 93.25 & 78.06 & 77.22 \\
 & socialbench\_sap\_en & 75.92 & 90.97 & 91.66 & 42.54 & 83.31 & 86.05 & 93.71 & 85.91 & 85.77 \\
 & socialbench\_sa\_rolestyle\_zh & 37.15 & 70.28 & 51.08 & 26.01 & 51.39 & 63.47 & 76.78 & 64.40 & 55.11 \\
 & socialbench\_sa\_rolestyle\_en & 47.16 & 66.35 & 57.97 & 40.68 & 72.30 & 81.49 & 63.65 & 84.73 & 77.70 \\
 & socialbench\_sa\_roleknowledge\_zh & 55.56 & 80.00 & 74.32 & 39.26 & 69.63 & 80.49 & 92.59 & 79.75 & 65.93 \\
 & socialbench\_sa\_roleknowledge\_en & 55.87 & 80.57 & 77.73 & 43.52 & 73.38 & 86.64 & 76.92 & 83.91 & 81.68 \\
 & socialbench\_mem\_short\_zh & 46.19 & 83.01 & 85.97 & 47.42 & 59.52 & 81.56 & 92.47 & 85.70 & 85.38 \\
 & socialbench\_mem\_short\_en & 55.32 & 68.97 & 45.39 & 33.98 & 41.83 & 82.03 & 88.56 & 76.20 & 82.26 \\
 & socialbench\_mem\_long\_zh & 57.68 & 79.25 & 80.45 & 44.59 & 65.73 & 88.82 & 91.50 & 88.60 & 86.75 \\
 & socialbench\_mem\_long\_en & 59.62 & 68.01 & 44.13 & 32.28 & 65.96 & 89.14 & 88.87 & 82.26 & 80.96 \\
 \midrule
\multirow{4}{*}{InCharacter} & incharacter\_16Personalities\_en & 0.58 & 0.45 & 0.54 & 0.54 & 0.54 & 0.44 & 0.44 & 0.47 & 0.46 \\
 & incharacter\_16Personalities\_zh & 0.36 & 0.36 & 0.68 & 0.55 & 0.50 & 0.68 & 0.77 & 0.64 & 0.68 \\
 & incharacter\_BFI\_en & 0.50 & 0.52 & 0.50 & 0.39 & 0.54 & 0.54 & 0.50 & 0.50 & 0.45 \\
 & incharacter\_BFI\_zh & 0.50 & 0.57 & 0.57 & 0.57 & 0.53 & 0.47 & 0.67 & 0.57 & 0.63 \\
 \midrule
Charactereval & charactereval & 2.23 & 2.49 & 2.53 & 2.14 & 2.24 & 2.89 & 2.90 & 2.63 & 2.26 \\
 \midrule
\multirow{3}{*}{RoleBench} & rolebench\_instruction\_generalization\_en & 20.13 & 20.73 & 21.95 & 13.28 & 15.72 & 16.87 & 17.93 & 25.39 & 18.89 \\
 & rolebench\_instruction\_generalization\_zh & 15.89 & 17.68 & 17.32 & 9.72 & 15.95 & 17.54 & 18.81 & 20.90 & 14.38 \\
 & rolebench\_generalization\_en & 20.66 & 20.70 & 21.72 & 12.57 & 14.22 & 15.29 & 15.60 & 24.55 & 17.25 \\
\bottomrule
\end{tabular}%
}
\caption{Direct zero-shot role-playing performance of different LLMs on various role-playing benchmarks (part2).}
\label{tab:zero_shot_performance_part_2}
\end{table*}

\section{Detailed Metrics}\label{app:sec:metrics}

\begin{itemize}
    \item \textbf{Accuracy}: Used for multiple-choice evaluation in Cross-MR and the SocialBench multiple-choice task.
    \item \textbf{ROUGE}: Used for text generation evaluation in HPD and RoleBench.
    \item \textbf{Exact Match}: Used in the SocialBench Generation Task to assess the alignment of key memory points.
    \item \textbf{LLM-as-a-Judge (Prompting)}: Used in InCharacter to convert model responses into standardized answers through analytical transformation.
    \item \textbf{LLM-as-a-Judge (Reward Model)}: Used in CharacterEval, where a reward model assigns scores across 12 evaluation dimensions for model responses.
\end{itemize}

\section{Detailed Results on Role-palying Benchmarks}\label{app:sec:results}

Tables~\ref{tab:zero_shot_performance_part_1}, \ref{tab:zero_shot_performance_part_2}, \ref{tab:cot_performance_part_1}, \ref{tab:cot_performance_part_2}, and \ref{tab:o1_performance} present the experimental results of various models using three role-playing methods across six benchmarks.

\begin{table*}[ht]
\centering
\resizebox{\textwidth}{!}{%
\begin{tabular}{llcccccccc}
\toprule
\textbf{Dataset} & \textbf{Sub Dataset} & \textbf{\makecell{OpenAI \\ GPT-4 Turbo}} & \textbf{\makecell{Qwen2.5-0.5B\\-Instruct}} & \textbf{\makecell{Qwen2.5-1.5B\\-Instruct}} & \textbf{\makecell{Qwen2.5\\-3B-Instruct}} & \textbf{\makecell{Qwen2.5-7B\\-Instruct}} & \textbf{\makecell{Qwen2.5-14B\\-Instruct}} & \textbf{\makecell{Qwen2.5-32B\\-Instruct}} & \textbf{\makecell{Qwen2.5-72B\\-Instruct}} \\
\midrule
\multirow{1}{*}{COT} & cross & 46.74 & 26.97 & 32.13 & 39.78 & 41.80 & 58.20 & 48.54 & 48.76 \\
\midrule
\multirow{2}{*}{HPD} & hpd\_en & 15.37 & 7.88 & 11.38 & 11.45 & 15.01 & 14.82 & 15.36 & 10.03 \\
 & hpd\_zh & 20.75 & 13.49 & 13.29 & 15.73 & 19.44 & 20.66 & 21.73 & 20.75 \\
 \midrule
\multirow{10}{*}{SocialBench} & socialbench\_sap\_zh & 70.21 & 32.32 & 64.22 & 75.27 & 77.22 & 75.02 & 82.53 & 75.95 \\
 & socialbench\_sap\_en & 59.10 & 29.55 & 59.78 & 78.25 & 85.64 & 87.41 & 87.55 & 84.13 \\
 & socialbench\_sa\_rolestyle\_zh & 54.18 & 26.63 & 45.82 & 51.39 & 69.66 & 70.28 & 78.02 & 72.45 \\
 & socialbench\_sa\_rolestyle\_en & 50.41 & 30.27 & 46.49 & 62.70 & 63.38 & 80.27 & 85.14 & 81.76 \\
 & socialbench\_sa\_roleknowledge\_zh & 66.67 & 32.35 & 61.98 & 83.70 & 89.38 & 95.06 & 94.07 & 97.53 \\
 & socialbench\_sa\_roleknowledge\_en & 46.05 & 37.25 & 67.61 & 82.39 & 89.47 & 95.04 & 95.75 & 93.52 \\
 & socialbench\_mem\_short\_zh & 79.09 & 13.17 & 28.98 & 72.04 & 77.69 & 87.58 & 82.42 & 83.98 \\
 & socialbench\_mem\_short\_en & 76.75 & 22.49 & 40.18 & 68.91 & 78.75 & 86.48 & 84.26 & 85.51 \\
 & socialbench\_mem\_long\_zh & 39.74 & 12.16 & 38.22 & 67.98 & 78.20 & 86.69 & 84.07 & 83.79 \\
 & socialbench\_mem\_long\_en & 57.18 & 14.44 & 54.29 & 67.21 & 79.66 & 86.98 & 82.15 & 84.65 \\
 \midrule
\multirow{4}{*}{InCharacter} & incharacter\_16Personalities\_en & 0.75 & 0.49 & 0.47 & 0.49 & 0.54 & 0.43 & 0.52 & 0.53 \\
 & incharacter\_16Personalities\_zh & 0.65 & 0.68 & 0.50 & 0.68 & 0.36 & 0.68 & 0.55 & 0.50 \\
 & incharacter\_BFI\_en & 0.40 & 0.43 & 0.50 & 0.49 & 0.52 & 0.49 & 0.49 & 0.42 \\
 & incharacter\_BFI\_zh & 0.73 & 0.63 & 0.67 & 0.60 & 0.63 & 0.53 & 0.70 & 0.70 \\
 \midrule
Charactereval & charactereval & 2.58 & 1.81 & 2.52 & 2.39 & 2.65 & 2.99 & 3.04 & 3.10 \\
\midrule
\multirow{3}{*}{RoleBench} & instruction\_generalization\_en & 27.13 & 16.31 & 22.33 & 22.66 & 25.81 & 25.55 & 24.60 & 23.61 \\
 & instruction\_generalization\_zh & 21.19 & 14.12 & 17.82 & 16.55 & 19.83 & 18.95 & 19.81 & 18.38 \\
 & role\_generalization\_en & 27.42 & 18.04 & 22.86 & 22.73 & 25.99 & 25.77 & 25.24 & 24.42 \\
\bottomrule
\end{tabular}%
}
\caption{Performance of role-playing with chain-of-thought reasoning on various role-playing benchmarks (part 1).}
\label{tab:cot_performance_part_1}
\end{table*}

\begin{table*}[ht]
\centering
\resizebox{\textwidth}{!}{%
\begin{tabular}{llccccccccc}
\toprule
\textbf{Dataset} & \textbf{Sub Dataset} & \textbf{\makecell{Gemma-2\\-2B-it}} & \textbf{\makecell{Gemma-2\\-9B-it}} & \textbf{\makecell{Gemma-2\\-27B-it}} & \textbf{\makecell{Llama-3.2-1B\\-Instruct}} & \textbf{\makecell{Llama-3.2-3B\\-Instruct}} & \textbf{\makecell{Llama-3.1-8B\\-Instruct}} & \textbf{\makecell{Llama-3.3-70B\\-Instruct}} & \textbf{\makecell{Mistral-8B\\-Instruct-2410}} & \textbf{\makecell{Mistral-7B\\-Instruct-v0.3}} \\
\midrule
Cross & cross & 33.48 & 50.11 & 42.25 & 27.87 & 40.45 & 46.97 & 62.70 & 39.10 & 41.12 \\
\midrule
\multirow{2}{*}{HPD} & hpd\_en & 7.01 & 8.04 & 7.39 & 6.84 & 11.17 & 4.90 & 6.85 & 15.15 & 10.68 \\
 & hpd\_zh & 13.49 & 14.82 & 9.81 & 12.48 & 16.04 & 16.60 & 20.27 & 12.82 & 15.64 \\
 \midrule
\multirow{10}{*}{SocialBench} & socialbench\_sap\_zh & 71.31 & 85.91 & 85.74 & 33.42 & 78.99 & 85.65 & 92.07 & 80.25 & 74.60 \\
 & socialbench\_sap\_en & 77.15 & 88.37 & 91.93 & 41.04 & 80.44 & 66.21 & 95.21 & 63.34 & 58.69 \\
 & socialbench\_sa\_rolestyle\_zh & 49.85 & 68.73 & 69.04 & 23.22 & 50.77 & 61.30 & 77.09 & 66.87 & 52.94 \\
 & socialbench\_sa\_rolestyle\_en & 64.19 & 80.41 & 82.70 & 30.95 & 55.14 & 54.46 & 89.19 & 48.65 & 50.95 \\
 & socialbench\_sa\_roleknowledge\_zh & 69.63 & 85.68 & 89.88 & 24.94 & 64.94 & 78.77 & 92.35 & 77.78 & 68.15 \\
 & socialbench\_sa\_roleknowledge\_en & 74.19 & 93.52 & 93.42 & 33.60 & 71.76 & 70.85 & 95.34 & 47.98 & 44.64 \\
 & socialbench\_mem\_short\_zh & 63.80 & 81.29 & 82.42 & 31.99 & 45.39 & 81.13 & 91.67 & 90.00 & 71.40 \\
 & socialbench\_mem\_short\_en & 70.49 & 70.57 & 49.78 & 27.71 & 50.84 & 85.37 & 89.23 & 85.54 & 86.62 \\
 & socialbench\_mem\_long\_zh & 63.67 & 83.56 & 82.16 & 29.47 & 35.65 & 74.90 & 89.66 & 87.70 & 73.89 \\
 & socialbench\_mem\_long\_en & 63.52 & 75.21 & 52.53 & 28.60 & 60.87 & 86.10 & 87.19 & 86.11 & 83.60 \\
 \midrule
\multirow{4}{*}{InCharacter} & incharacter\_16Personalities\_en & 0.56 & 0.51 & 0.55 & 0.48 & 0.54 & 0.49 & 0.51 & 0.44 & 0.53 \\
 & incharacter\_16Personalities\_zh & 0.59 & 0.68 & 0.55 & 0.64 & 0.64 & 0.73 & 0.77 & 0.45 & 0.45 \\
 & incharacter\_BFI\_en & 0.49 & 0.47 & 0.48 & 0.55 & 0.49 & 0.47 & 0.54 & 0.46 & 0.52 \\
 & incharacter\_BFI\_zh & 0.57 & 0.70 & 0.53 & 0.67 & 0.67 & 0.70 & 0.63 & 0.53 & 0.60 \\
 \midrule
Charactereval & charactereval & 2.29 & 2.48 & 2.56 & 2.15 & 2.11 & 2.67 & 2.70 & 2.55 & 2.21 \\
\midrule
\multirow{3}{*}{RoleBench} & instruction\_generalization\_en & 22.18 & 23.54 & 23.55 & 17.53 & 19.48 & 22.38 & 20.72 & 28.03 & 24.89 \\
 & instruction\_generalization\_zh & 16.31 & 18.67 & 18.65 & 14.58 & 16.03 & 19.38 & 19.58 & 20.83 & 18.04 \\
 & role\_generalization\_en & 23.52 & 24.33 & 24.73 & 15.59 & 17.17 & 21.74 & 17.79 & 26.70 & 23.00 \\
\bottomrule
\end{tabular}%
}
\caption{Performance of role-playing with chain-of-thought reasoning on various role-playing benchmarks (part 2).}
\label{tab:cot_performance_part_2}
\end{table*}

\begin{table*}[ht]
\centering
\resizebox{\textwidth}{!}{%
\begin{tabular}{llccccccc}
\toprule
\textbf{Dataset} & \textbf{Sub Dataset} & \textbf{\makecell{OpenAI\\o1-mini}} & \textbf{\makecell{Deepseek\\-R1}} & \textbf{\makecell{QwQ-32B\\-Preview}} & \textbf{\makecell{DeepSeek-R1\\-Distill-Llama-8B}} & \textbf{\makecell{DeepSeek-R1\\-Distill-Qwen-14B}} & \textbf{\makecell{DeepSeek-R1\\-Distill-Qwen-32B}} & \textbf{\makecell{DeepSeek-R1\\-Distill-Llama-70B}} \\
\midrule
Cross & cross & 45.23 & 44.19 & 62.02 & 45.17 & 46.97 & 50.11 & 51.91 \\
\midrule
\multirow{2}{*}{HPD} & hpd\_en & 14.96 & 13.28 & 3.95 & 9.25 & 12.21 & 12.41 & 12.04 \\
 & hpd\_zh & 19.87 & 20.87 & 10.14 & 9.22 & 17.48 & 18.58 & 15.85 \\
 \midrule
\multirow{10}{*}{SocialBench} & socialbench\_sap\_zh & 68.11 & 64.72 & 79.16 & 77.97 & 81.94 & 88.69 & 89.70 \\
 & socialbench\_sap\_en & 57.08 & 56.52 & 85.09 & 83.31 & 88.92 & 81.67 & 94.12 \\
 & socialbench\_sa\_rolestyle\_zh & 52.29 & 56.87 & 70.90 & 64.09 & 72.76 & 78.02 & 78.02 \\
 & socialbench\_sa\_rolestyle\_en & 48.29 & 46.33 & 84.05 & 67.97 & 77.97 & 61.22 & 87.84 \\
 & socialbench\_sa\_roleknowledge\_zh & 64.89 & 62.10 & 94.81 & 81.23 & 93.33 & 93.58 & 94.57 \\
 & socialbench\_sa\_roleknowledge\_en & 44.17 & 47.21 & 95.55 & 85.83 & 82.89 & 80.47 & 97.06 \\
 & socialbench\_mem\_short\_zh & 77.42 & 77.82 & 83.09 & 71.77 & 76.03 & 80.91 & 86.88 \\
 & socialbench\_mem\_short\_en & 74.23 & 73.28 & 75.68 & 78.60 & 80.13 & 89.87 & 90.47 \\
 & socialbench\_mem\_long\_zh & 37.42 & 39.26 & 82.03 & 73.34 & 79.87 & 85.45 & 86.80 \\
 & socialbench\_mem\_long\_en & 55.71 & 57.45 & 71.46 & 80.40 & 84.14 & 88.69 & 89.29 \\
 \midrule
\multirow{4}{*}{InCharacter} & incharacter\_16Personalities\_en & 0.67 & 0.67 & 0.51 & 0.52 & 0.47 & 0.61 & 0.60 \\
 & incharacter\_16Personalities\_zh & 0.52 & 0.59 & 0.45 & 0.64 & 0.59 & 0.55 & 0.64 \\
 & incharacter\_BFI\_en & 0.44 & 0.41 & 0.49 & 0.50 & 0.49 & 0.41 & 0.50 \\
 & incharacter\_BFI\_zh & 0.63 & 0.63 & 0.43 & 0.53 & 0.50 & 0.57 & 0.43 \\
 \midrule
Charactereval & charactereval & 2.56 & 2.32 & 2.78 & 2.38 & 2.70 & 2.69 & 2.63 \\
\midrule
\multirow{3}{*}{RoleBench} & instruction\_generalization\_en & 24.91 & 23.64 & 9.54 & 19.54 & 20.45 & 17.66 & 19.54 \\
 & instruction\_generalization\_zh & 20.29 & 18.73 & 17.45 & 12.22 & 18.31 & 15.86 & 14.73 \\
 & role\_generalization\_en & 26.66 & 24.47 & 10.97 & 18.21 & 19.76 & 16.38 & 17.66 \\
\bottomrule
\end{tabular}%
}
\caption{Performance of role-playing using reasoning-optimized LLMs on various role-playing benchmarks.}
\label{tab:o1_performance}
\end{table*}

\section{Fine-Grained Results on CharacterEval}\label{app:sec:charactereval}

Table~\ref{tab:charactereval_zero_shot_performance_1}, \ref{tab:charactereval_zero_shot_performance_2}, \ref{tab:charactereval_cot_performance_1}, \ref{tab:charactereval_cot_performance_2}, and \ref{tab:charactereval_o1_performance} present the fine-grained performance of different models on CharacterEval using three role-playing methods.
The evaluation consists of 12 dimensions, with all scores provided by a pretrained reward model.

\begin{table*}[!htb]
\centering
\resizebox{\textwidth}{!}{%
\begin{tabular}{lcccccccc}
\toprule
\textbf{Metric} & \textbf{\makecell{OpenAI \\ GPT-4 Turbo}} & \textbf{\makecell{Qwen2.5-0.5B\\-Instruct}} & \textbf{\makecell{Qwen2.5-1.5B\\-Instruct}} & \textbf{\makecell{Qwen2.5\\-3B-Instruct}} & \textbf{\makecell{Qwen2.5-7B\\-Instruct}} & \textbf{\makecell{Qwen2.5-14B\\-Instruct}} & \textbf{\makecell{Qwen2.5-32B\\-Instruct}} & \textbf{\makecell{Qwen2.5-72B\\-Instruct}} \\
\midrule
Accuracy & 2.92  & 1.90  & 2.35  & 2.86  & 2.98  & 3.17  & 3.10  & 3.29  \\
Behavior & 1.39  & 1.29  & 1.19  & 1.26  & 1.67  & 2.40  & 2.20  & 2.65  \\
Coherence & 3.29  & 2.20  & 2.76  & 3.19  & 3.60  & 3.90  & 3.76  & 3.93  \\
Communication Skills & 2.68  & 1.66  & 1.87  & 2.75  & 2.98  & 3.25  & 3.15  & 3.53  \\
Consistency & 2.90  & 1.83  & 2.48  & 2.73  & 3.27  & 3.65  & 3.46  & 3.64  \\
Diversity & 1.36  & 1.19  & 1.19  & 1.31  & 1.60  & 2.00  & 1.93  & 2.18  \\
Empathy & 2.94  & 1.88  & 2.33  & 2.86  & 3.14  & 3.33  & 3.21  & 3.46  \\
Exposure & 1.91  & 1.39  & 1.42  & 2.02  & 2.03  & 2.18  & 2.21  & 2.47  \\
Fluency & 3.03  & 1.94  & 2.42  & 2.90  & 3.34  & 3.53  & 3.39  & 3.63  \\
Hallucination & 2.55  & 1.70  & 2.08  & 2.46  & 2.78  & 3.05  & 2.94  & 3.17  \\
Humanlikeness & 2.78  & 1.75  & 2.41  & 2.59  & 3.11  & 3.48  & 3.30  & 3.47  \\
Utterance & 2.48  & 1.67  & 2.11  & 2.41  & 2.74  & 3.04  & 2.90  & 3.09  \\
\bottomrule
\end{tabular}%
}
\caption{Fine-grained results of direct zero-shot role-playing on CharacterEval benchmark (part1).}
\label{tab:charactereval_zero_shot_performance_1}
\end{table*}

\begin{table*}[ht]
\centering
\resizebox{\textwidth}{!}{%
\begin{tabular}{lccccccccc}
\toprule
\textbf{Metrics}  & \textbf{\makecell{Gemma-2\\-2B-it}} & \textbf{\makecell{Gemma-2\\-9B-it}} & \textbf{\makecell{Gemma-2\\-27B-it}} & \textbf{\makecell{Llama-3.2-1B\\-Instruct}} & \textbf{\makecell{Llama-3.2-3B\\-Instruct}} & \textbf{\makecell{Llama-3.1-8B\\-Instruct}} & \textbf{\makecell{Llama-3.3-70B\\-Instruct}} & \textbf{\makecell{Mistral-8B\\-Instruct-2410}} & \textbf{\makecell{Mistral-7B\\-Instruct-v0.3}} \\
\midrule
Accuracy & 2.52 & 2.80 & 2.90 & 2.37 & 2.56 & 2.94 & 3.12 & 2.85 & 2.62  \\
Behavior & 1.20 & 1.14 & 1.16 & 1.86 & 1.58 & 2.61 & 2.04 & 1.16 & 1.22  \\
Coherence & 3.00 & 3.42 & 3.47 & 2.69 & 2.87 & 3.55 & 3.70 & 3.65 & 3.04  \\
Communication Skills & 2.16 & 2.12 & 2.10 & 2.08 & 2.15 & 2.94 & 2.99 & 2.61 & 2.53  \\
Consistency & 2.67 & 3.27 & 3.38 & 2.23 & 2.53 & 3.31 & 3.43 & 3.36 & 2.49  \\
Diversity & 1.25 & 1.20 & 1.25 & 1.57 & 1.47 & 2.18 & 1.85 & 1.22 & 1.26  \\
Empathy & 2.51 & 2.78 & 2.81 & 2.36 & 2.49 & 3.02 & 3.16 & 3.06 & 2.64  \\
Exposure & 1.59 & 1.50 & 1.51 & 1.67 & 1.67 & 2.12 & 2.06 & 1.79 & 1.91  \\
Fluency & 2.74 & 3.09 & 3.10 & 2.44 & 2.60 & 3.28 & 3.40 & 3.30 & 2.68  \\
Hallucination & 2.23 & 2.56 & 2.55 & 2.05 & 2.17 & 2.73 & 2.87 & 2.69 & 2.26  \\
Humanlikeness & 2.64 & 3.34 & 3.43 & 2.30 & 2.58 & 3.18 & 3.31 & 3.20 & 2.29  \\ 
Utterance & 2.26 & 2.61 & 2.67 & 2.09 & 2.20 & 2.77 & 2.92 & 2.71 & 2.20  \\
\bottomrule
\end{tabular}%
}
\caption{Fine-grained results of direct zero-shot role-playing on CharacterEval benchmark (part2).}
\label{tab:charactereval_zero_shot_performance_2}
\end{table*}

\begin{table*}[ht]
\centering
\resizebox{\textwidth}{!}{%
\begin{tabular}{lcccccccc}
\toprule
\textbf{Metric} & \textbf{\makecell{OpenAI \\ GPT-4 Turbo}} & \textbf{\makecell{Qwen2.5 \\ 0.5B-Instruct}} & \textbf{\makecell{Qwen2.5 \\ 1.5B-Instruct}} & \textbf{\makecell{Qwen2.5 \\ 3B-Instruct}} & \textbf{\makecell{Qwen2.5 \\ 7B-Instruct}} & \textbf{\makecell{Qwen2.5 \\ 14B-Instruct}} & \textbf{\makecell{Qwen2.5 \\ 32B-Instruct}} & \textbf{\makecell{Qwen2.5 \\ 72B-Instruct}} \\
\midrule
{Accuracy} & 2.87 & 2.13 & 2.76 & 2.81 & 2.96 & 3.17 & 3.19 & 3.26  \\
{Behavior} & 2.00 & 1.37 & 1.52 & 1.30 & 1.53 & 2.18 & 2.38 & 2.25  \\
{Coherence} & 3.27 & 2.28 & 3.41 & 3.14 & 3.49 & 3.76 & 3.79 & 3.84  \\
{Communication Skills} & 2.98 & 1.73 & 1.72 & 2.67 & 3.00 & 3.23 & 3.26 & 3.51  \\
{Consistency} & 2.72 & 1.86 & 3.35 & 2.63 & 3.06 & 3.48 & 3.49 & 3.52  \\
{Diversity} & 1.70 & 1.31 & 1.43 & 1.28 & 1.47 & 1.89 & 2.04 & 2.01  \\
{Empathy} & 2.93 & 1.96 & 2.71 & 2.82 & 3.03 & 3.29 & 3.32 & 3.44  \\
{Exposure} & 2.26 & 1.61 & 1.39 & 2.03 & 2.11 & 2.21 & 2.27 & 2.50  \\
{Fluency} & 2.90 & 2.04 & 3.04 & 2.82 & 3.09 & 3.45 & 3.49 & 3.50  \\
{Hallucination} & 2.41 & 1.79 & 2.50 & 2.39 & 2.66 & 2.95 & 2.98 & 3.06  \\
{Humanlikeness} & 2.51 & 1.87 & 3.69 & 2.48 & 2.75 & 3.36 & 3.32 & 3.28  \\
{Utterance} & 2.44 & 1.73 & 2.73 & 2.32 & 2.62 & 2.92 & 2.93 & 2.99  \\
\bottomrule
\end{tabular}%
}
\caption{Fine-grained results of role-playing with CoT on CharacterEval benchmark (part 1).}
\label{tab:charactereval_cot_performance_1}
\end{table*}

\begin{table*}[ht]
\centering
\resizebox{\textwidth}{!}{%
\begin{tabular}{lccccccccc}
\toprule
\textbf{Metrics}  & \textbf{\makecell{Gemma-2\\-2B-it}} & \textbf{\makecell{Gemma-2\\-9B-it}} & \textbf{\makecell{Gemma-2\\-27B-it}} & \textbf{\makecell{Llama-3.2-1B\\-Instruct}} & \textbf{\makecell{Llama-3.2-3B\\-Instruct}} & \textbf{\makecell{Llama-3.1-8B\\-Instruct}} & \textbf{\makecell{Llama-3.3-70B\\-Instruct}} & \textbf{\makecell{Mistral-8B\\-Instruct-2410}} & \textbf{\makecell{Mistral-7B\\-Instruct-v0.3}} \\
\midrule
{Accuracy} & 2.63 & 2.88 & 2.91 & 2.52 & 2.51 & 2.90 & 3.07 & 2.87 & 2.57  \\
{Behavior} & 1.28 & 1.11 & 1.14 & 1.66 & 1.20 & 2.01 & 1.34 & 1.23 & 1.39  \\
{Coherence} & 3.11 & 3.38 & 3.50 & 2.71 & 2.79 & 3.38 & 3.58 & 3.43 & 2.86  \\
{Communication Skills} & 1.88 & 2.23 & 2.33 & 2.06 & 1.96 & 2.84 & 2.91 & 2.71 & 2.39  \\
{Consistency} & 2.90 & 3.19 & 3.30 & 2.22 & 2.46 & 2.98 & 3.20 & 3.03 & 2.37  \\
{Diversity} & 1.29 & 1.18 & 1.20 & 1.50 & 1.22 & 1.76 & 1.35 & 1.26 & 1.35  \\
{Empathy} & 2.53 & 2.77 & 2.92 & 2.38 & 2.41 & 2.95 & 3.08 & 3.01 & 2.55  \\
{Exposure} & 1.44 & 1.59 & 1.58 & 1.77 & 1.52 & 2.04 & 2.01 & 1.93 & 1.83  \\
{Fluency} & 2.71 & 3.00 & 3.20 & 2.47 & 2.51 & 3.06 & 3.30 & 3.13 & 2.59  \\
{Hallucination} & 2.21 & 2.53 & 2.60 & 2.13 & 2.09 & 2.63 & 2.76 & 2.57 & 2.15  \\
{Humanlikeness} & 3.06 & 3.30 & 3.35 & 2.28 & 2.54 & 2.89 & 3.03 & 2.87 & 2.32  \\ 
{Utterance} & 2.40 & 2.58 & 2.66 & 2.12 & 2.13 & 2.60 & 2.72 & 2.52 & 2.18  \\
\bottomrule
\end{tabular}%
}
\caption{Fine-grained results of role-playing with CoT on CharacterEval benchmark (part 2).}
\label{tab:charactereval_cot_performance_2}
\end{table*}

\begin{table*}[ht]
\centering
\resizebox{\textwidth}{!}{
\begin{tabular}{lccccccc}
\toprule
\textbf{Metrics} & \textbf{\makecell{OpenAI\\o1-mini}} & \textbf{\makecell{Deepseek\\-R1}} & \textbf{\makecell{QwQ-32B\\-Preview}} & \textbf{\makecell{DeepSeek-R1\\-Distill-Llama-8B}} & \textbf{\makecell{DeepSeek-R1\\-Distill-Qwen-14B}} & \textbf{\makecell{DeepSeek-R1\\-Distill-Qwen-32B}} & \textbf{\makecell{DeepSeek-R1\\-Distill-Llama-70B}} \\
\midrule
Accuracy & 2.45 & 2.38 & 3.08 & 2.65 & 2.95 & 3.06 & 2.96 \\
Behavior & 1.29 & 1.27 & 1.64 & 1.63 & 1.62 & 1.31 & 1.43 \\
Coherence & 3.61 & 3.19 & 3.61 & 3.08 & 3.46 & 3.55 & 3.47 \\
Communication Skills & 2.42 & 2.31 & 2.94 & 2.40 & 2.93 & 2.83 & 2.79 \\
Consistency & 2.87 & 3.02 & 3.31 & 2.68 & 3.15 & 3.25 & 3.12 \\
Diversity & 1.19 & 1.20 & 1.53 & 1.49 & 1.52 & 1.33 & 1.40 \\
Empathy & 2.98 & 3.21 & 3.14 & 2.61 & 3.03 & 3.08 & 2.97 \\
Exposure & 1.94 & 2.08 & 2.09 & 1.88 & 2.06 & 2.01 & 1.99 \\
Fluency & 3.44 & 2.29 & 3.30 & 2.77 & 3.23 & 3.23 & 3.10 \\
Hallucination & 2.81 & 2.11 & 2.85 & 2.35 & 2.76 & 2.78 & 2.68 \\
Humanlikeness & 3.21 & 2.69 & 3.15 & 2.68 & 3.01 & 3.09 & 2.99 \\
Utterance & 2.54 & 2.12 & 2.76 & 2.36 & 2.66 & 2.71 & 2.62 \\
\bottomrule
\end{tabular}
}
\caption{Fine-grained results of various reasoning-enhanced models on CharacterEval.}
\label{tab:charactereval_o1_performance}
\end{table*}

\section{Limitations of Reinforcement Learning for Role-Playing Tasks}\label{app:sec:rllimit}

To address the potential trade-offs between reasoning-optimized models and role-playing tasks, we conducted a focused analysis using DeepSeek-R1 as a representative case, leveraging its publicly available technical details. 
DeepSeek-R1-Zero is trained exclusively through reinforcement learning, without supervised fine-tuning, employing a rule-based reward system designed to enhance reasoning accuracy and structural compliance. 
While this training paradigm achieves substantial gains in formal reasoning tasks, it introduces several critical limitations that hinder performance in persona-grounded interactions. 
First, the \textbf{reasoning–role conflict} arises because the model is explicitly optimized for factual correctness and logical precision, often at the expense of the creativity, empathy, and subjectivity necessary for maintaining in-character responses. 
Second, \textbf{style drift} emerges as the structured reasoning formats (e.g., encapsulated reasoning tags) promote rigid and formal outputs, undermining the expressive and stylistically diverse language required for immersive role-play. 
Third, \textbf{reward misalignment} is evident, as the reinforcement signals focus narrowly on deterministic correctness and formatting adherence, neglecting essential dimensions such as persona consistency, emotional tone, and conversational adaptability. 
Finally, the model exhibits \textbf{over-optimization for test-time computation}, favoring extended reasoning chains and reflection loops (e.g., “aha moments”) that, while useful for complex problem-solving, lead to verbose, unnatural, and character-inconsistent outputs during role-play. 
These observations highlight a fundamental misalignment between reasoning-focused optimization and the adaptive, expressive requirements of role-playing, offering important insights for future efforts to harmonize reasoning capabilities with character fidelity in large language models.

\end{document}